\documentclass{article}
\usepackage{arxiv}
\usepackage{hyperref}
\usepackage{microtype}

\usepackage{amssymb}
\usepackage{amsmath}
\usepackage{amsthm}
\usepackage{mathdots}
\usepackage{mathtools}
\usepackage{xypic}
\usepackage{multicol}

\usepackage{tensor}
\usepackage{sectsty}

\usepackage{urwchancal}
\DeclareFontFamily{OT1}{pzc}{}
\DeclareFontShape{OT1}{pzc}{m}{it}{<-> s * [1.10] pzcmi7t}{}
\DeclareMathAlphabet{\mathpzc}{OT1}{pzc}{m}{it}

\usepackage{graphicx}
\usepackage{ifpdf}
\ifpdf
\usepackage{epstopdf}
\PrependGraphicsExtensions{.svg}
\fi

\usepackage{xypic}
\usepackage{booktabs}
\usepackage{subfigure}
\usepackage{array,multirow}
\usepackage{rotating}
\usepackage{makecell}
\usepackage{rotating}
\usepackage{array}
\usepackage{float}
\restylefloat{table}
\usepackage{physics}
\usepackage{algorithm}
\usepackage{algpseudocode}
\usepackage{bbding}

\usepackage{wrapfig}

\newcommand{\etal}{\textit{et al}. }
\newcommand{\ie}{\textit{i}.\textit{e}., }
\newcommand{\eg}{\textit{e}.\textit{g}., }

\newcommand{\publicationdetails}
{
	The paper has been accepted for publication at the Australian Conference on Robotics and Automation, 2019. 
}
\newcommand{\publicationversion}
{Author accepted version}

\begin{document}
\title{Event Camera Calibration of Per-pixel Biased Contrast Threshold}
\headertitle{Event Camera Calibration of Per-pixel Biased Contrast Threshold}

\author{
	\href{https://orcid.org/0000-0003-0815-1287}{\includegraphics[scale=0.06]{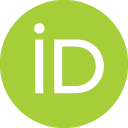}\hspace{1mm}
		Ziwei Wang}
	\\
	Systems Theory and Robotics Group \\
	Australian National University \\
	ACT, 2601, Australia \\
	\texttt{ziwei.wang1@anu.edu.au} \\
	\\
	\And
	\href{https://orcid.org/0000-0002-7764-298X}{\includegraphics[scale=0.06]{orcid.png}\hspace{1mm}
		Yonhon Ng}
	\\
	Systems Theory and Robotics Group \\
	Australian National University \\
	ACT, 2601, Australia \\
	\texttt{yonhon.ng@anu.edu.au} \\
	\\
	\And
	\href{https://orcid.org/0000-0003-4391-7014}{\includegraphics[scale=0.06]{orcid.png}\hspace{1mm}
		Pieter van Goor}
	\\
	Systems Theory and Robotics Group \\
	Australian National University \\
	ACT, 2601, Australia \\
	\texttt{pieter.vangoor@anu.edu.au} \\
	\And	\href{https://orcid.org/0000-0002-7803-2868}{\includegraphics[scale=0.06]{orcid.png}\hspace{1mm}
		Robert Mahony}
	\\
	Systems Theory and Robotics Group \\
	Australian National University \\
	ACT, 2601, Australia \\
	\texttt{robert.mahony@anu.edu.au} \\
}

\maketitle

\begin{abstract}
Event cameras output asynchronous events to represent intensity changes with a high temporal resolution, even under extreme lighting conditions. Currently, most of the existing works use a single contrast threshold to estimate the intensity change of all pixels.
However, complex circuit bias and manufacturing imperfections cause biased pixels and mismatch contrast threshold among pixels, which may lead to undesirable outputs. 
In this paper, we propose a new event camera model and two calibration approaches which cover event-only cameras and 
hybrid image-event cameras.
When intensity images are simultaneously provided along with events, we also propose an efficient online method to calibrate event cameras that adapts to time-varying event rates.
We demonstrate the advantages of our proposed methods compared to the state-of-the-art on several different event camera datasets.

\end{abstract}

\section{Introduction}
An event camera is a bio-inspired silicon retina that can provide high pixel bandwidth, high dynamic range and low latency measurement of image intensity changes~\cite{lichtsteiner2008128,posch2010qvga}.
In contrast to the traditional cameras, which generate intensity images at a fixed frequency, event cameras are based on the Dynamic Vision Sensor (DVS), which allows them to generate asynchronous events (ON or OFF signals) to represent changes of intensity. These events are generated by pixel-level brightness changes, which contain timestamps at microsecond resolution with a High Dynamic Range (HDR) \cite{Tedaldi2016feature,zhou2018semi}. 
The Dynamic and active pixel vision sensor (DAVIS) incorporates the DVS and a synchronous  frame-based  active  pixel  sensor (APS) \cite{berner2013240x180}, which also generates intensity images simultaneously. Another higher resolution ($480 \times 360$) event camera that can provide both grayscale intensity image along with events is the HVGA ATIS sensor~\cite{manderscheid2019speed}.

Due to the properties of an event camera, it is particularly suitable for applications where fast robotic vision is required. For example, an event camera was used in simultaneous localization and mapping (SLAM)~\cite{cook2011interacting,kim2008simultaneous,kim2016real,Vidal18}, feature detection~\cite{Tedaldi2016feature,scheerlinck2018computing}, feature tracking~\cite{gehrig2018asynchronous,Alzugaray2018asy}, visual odometry~\cite{Mueggler14,kueng2016low}, optical flow~\cite{bardow2016simultaneous}, and image reconstruction~\cite{pan2019bringing,orchard2014accelerated,scheerlinck2018computing}. 

In DVS, contrast threshold defines the logarithmic intensity change that will trigger an event. 
It is commonly assumed that the contrast threshold is a constant (typically 10\%) for all pixels \cite{scheerlinck2018computing,reinbacher2016real,kim2008simultaneous,kim2016real}.
However, due to a number of factors like event generation speed, intensity change, circuit noise, and manufacturing imperfections, the actual contrast threshold at each pixel may be different from the desired contrast threshold. In addition, undesirable junction leakage in the differencing amplifier and biased circuit cause noisy background positive events that are not correlated to the actual visual input~\cite{kueng2016low,yang2015dynamic,brandli2014real}. 

As previously reported, the effective contrast threshold can vary between 10-15\% \cite{serrano2013128,lichtsteiner2008128,berner2013240x180,lenero20113}. Algorithms that correctly model the variability of the inter-pixel contrast threshold and bias can achieve an improvement in the accuracy of their output. 

This paper presents novel calibration methods to correctly recover the different per-pixel contrast threshold for an event camera. 
The contribution of this paper are:
\begin{itemize}
	\item Analyse the source of contrast threshold variation at circuit level and introduce four types of pixels based on their different behaviours compared to the commonly used model.
	\item  Introduce a new event generation model that can represent inter-pixel biases and contrast thresholds.
	\item Propose both online and offline linear regression methods for event-only cameras (\eg DVS128 cameras \cite{lichtsteiner2008128}, DVS-Gen2 cameras \cite{son20174}) and event cameras with intensity images (\eg DAVIS and HVGA ATIS sensor).
\end{itemize}

The rest of this paper is organised as follow: we briefly introduce the prior works on estimating the contrast threshold in Section 2. In Section 3, we describe the traditional event generation model and our new model. We also propose four types of special pixels in this section. Section 4 describes our proposed methods. Then, Section 5 shows intensity image reconstruction results to verify our online and offline calibration methods. Section 6 concludes the paper.

\section{Related Work}

In the early design of event-based camera software \textit{jAER}, a filter was implemented to remove the background noisy positive events caused by transistor junction leakage or noise \cite{delbruck2008frame}. 
For each pixel, if no event happens at the neighbouring patch recently, the event that is detected at the pixel is potentially noise that does not correlated to actual intensity changes. The filter directly removes all isolated events without the support of their 8 neighbouring pixels within the previous period $T$, which largely removes the background noisy events \cite{delbruck2008frame}. 
The DAVIS camera and related software we used in our experiments have already suppressed the noise discussed in the paper, however, it still leaves distinct pixels that require special attention. 

Some researchers worked on improving the quality of the output of event cameras by increasing the threshold sensitivity, which would increase the intensity resolution and alleviate the quantization effects in fine texture reconstruction and recognition  \cite{brandli2014real,yang2015dynamic,serrano2013128,orchard2014accelerated}.
For example, Serrano \etal increased the threshold sensitivity from 10\% to 1.5\% by designing a novel transimpedance pre-amplification \cite{serrano2013128}.
Yang \etal also proposed an in-pixel asynchronous delta modulator (ADM) to replace the self-timed reset (STR) in event cameras, which increases the threshold sensitivity to 1\%. However, these designs still cannot solve the problem of threshold mismatch and bias \cite{yang2015dynamic}. Recently, Pan \etal obtained sharper reconstruction images by combining motion blurred intensity image and events. Their work computed an optimal (single) contrast threshold across every pixel with an energy function minimization method.  

Our work is most related to \cite{brandli2014real}, because they accounted for the variance of contrast threshold between pixels.
To improve the real-time video decompression performance, Brandli \etal estimated the unbalanced ON and OFF threshold and the mismatched threshold for each pixel by computing the direct integration error of events between two intensity image frames. The algorithm avoids cumulating the threshold error by continuously resetting ON and OFF threshold values for each pixel. 
However, the method still has drawbacks: 
1) The model lacks theoretical justification,
2) The method heavily relies on intensity images. Therefore the method cannot calibrate event cameras which can only generate events (\eg DVS128 cameras \cite{lichtsteiner2008128}, DVS-Gen2 cameras \cite{son20174}).

\section{Problem Formulation}

\subsection{Event Generation Model and Notation}

Consider a continuous image intensity function
\begin{gather}
\begin{aligned}
I: \; &\mathbb{N}^2 \times \mathbb{R} \to \mathbb{R^+} \\
& (x,y), t \mapsto I((x,y), t),
\end{aligned}
\end{gather}
where the inputs $(x,y)$ are some pixel coordinates, and $t$ is the time of measurement.
Define the continuous log image intensity function as $L := \log \circ I$.
Events are modelled as $e_i^{\vb*{p}}$ at given pixel coordinates $\vb*{p} = (\vb*{p}_x ,\vb*{p}_y)^{T}$ consisting of a timestamp $t_i^{\vb*{p}}$ and a polarity $\sigma^{\vb*{p}}_i$ such that
\begin{equation} \label{1.1}
e_i^{\vb*{p}} = (t_i^{\vb*{p}}, \sigma_i^{\vb*{p}}).
\end{equation}
The polarity $\sigma^{\vb*{p}}_i$ of an event $e_i^{\vb*{p}}$ is either $+1$ or $-1$, and we model the triggering of an event by
\begin{equation}  \label{1.4}
\sigma_i^{\vb*{p}} = 
\begin{cases}
+1, &  \Delta L^{\vb*{p}}(t^{\vb*{p}}_i) \geq c^{\vb*{p}}(t_i^{\vb*{p}}) + b^{\vb*{p}}(t_i^{\vb*{p}}),\\  
-1, &  \Delta L^{\vb*{p}}(t^{\vb*{p}}_i) \leq -c^{\vb*{p}}(t_i^{\vb*{p}}) + b^{\vb*{p}}(t_i^{\vb*{p}}),
\end{cases}  
\end{equation}  
where $c^{\vb*{p}}$ is the (time-varying) pixel-wise threshold value, $b^{\vb*{p}}$ is the (time-varying) pixel-wise bias value, and $\Delta L(t^{\vb*{p}}_i)$ is defined by
\begin{equation} \label{eq:delta-log}
\Delta L^{\vb*{p}}(t^{\vb*{p}}_i) = L^{\vb*{p}}(t^{\vb*{p}}_i) - L^{\vb*{p}}(t^{\vb*{p}}_{i-1}).
\end{equation}

\begin{figure}[t]
	\centering    
	\includegraphics[width=6.2cm]{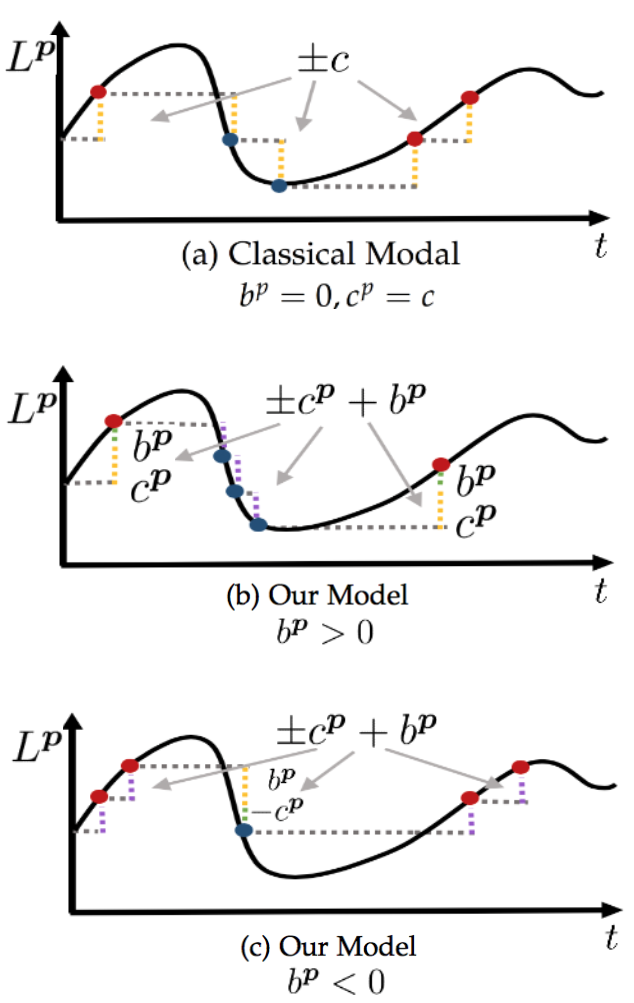}
	\caption{\label{image2} Event Generation Model. The classical model was used in \protect\cite{gehrig2019eklt,Tedaldi2016feature,scheerlinck2018computing}. 
	}
\end{figure}

Equation (\ref{1.1})  shows the proposed model for discrete time event signals, which allows for different biases and threshold values to apply to each pixel.
A new event $e^{\vb*{p}}_i = (t_i^{\vb*{p}},\sigma_i^{\vb*{p}})$ with positive polarity ($\sigma_i^{\vb*{p}} = +1$) is triggered when the log intensity changes $\Delta L^{\vb*{p}}(t^{\vb*{p}}_i) $ is larger than the biased contrast threshold at timestamp $t_i$, and vice versa.

\subsection{Special Pixels}
Most of the previous works model the contrast threshold as being consistent across all pixels in the event camera and ignore biased pixels altogether \cite{scheerlinck2018computing,reinbacher2016real,kim2008simultaneous,kim2016real}. This can be seen as a special case of our more general model in (\ref{1.2}) and can be written as

\begin{equation}  \label{1.2}
\sigma_i^{\vb*{p}} = \left\{  
\begin{array}{lr}  
+1, &  \Delta L^{\vb*{p}}(t_i^{\vb*{p}}) \geq c,\\  
-1, &  \Delta L^{\vb*{p}}(t_i^{\vb*{p}}) \leq -c,
\end{array}  
\right.  
\end{equation}  
where $\Delta L^{\vb*{p}}(t^{\vb*{p}}_i)$ is defined as in \eqref{eq:delta-log}.

\begin{figure*}[t]
	\centering    
	\includegraphics[width=17cm]{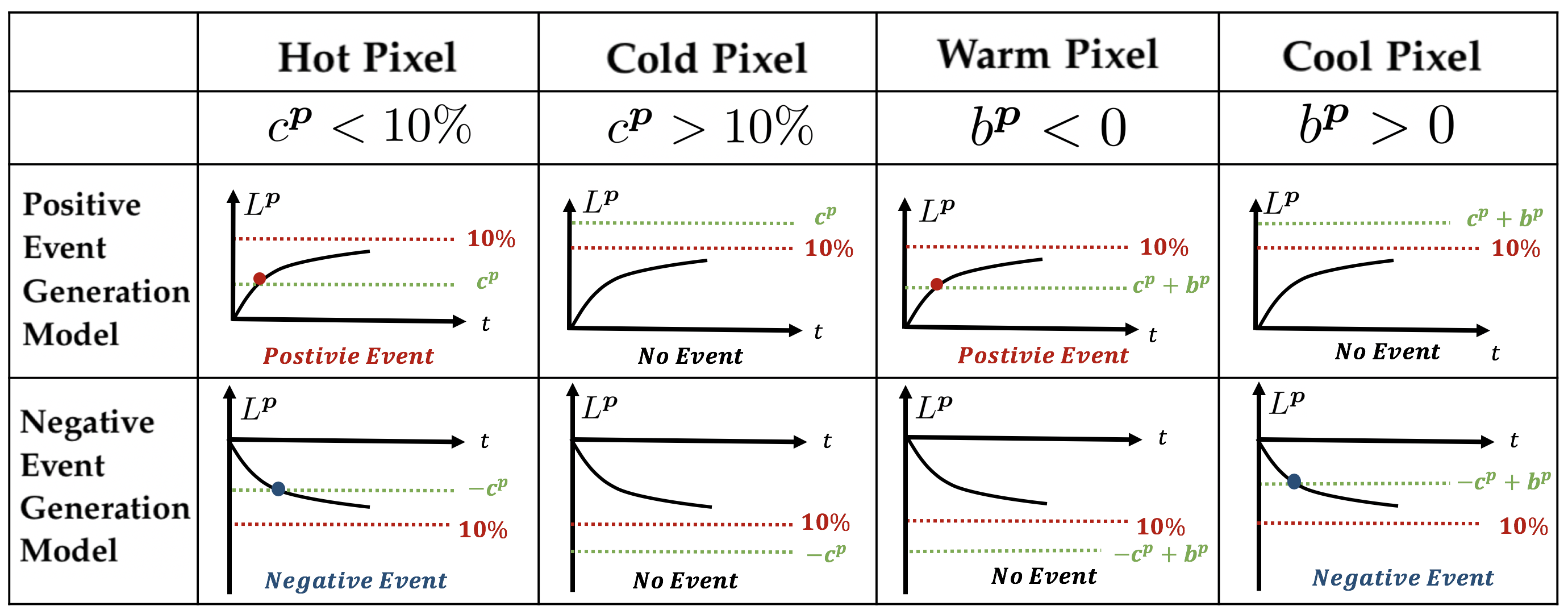}
	\caption{\label{fig:special_pixel} Four Types of Special Pixels}
\end{figure*}

If we ignore the mismatch in contrast threshold and the circuit bias, the classical model of event generation in Figure \ref{image2} will generate four types of special pixels, which will affect the performance of image reconstruction and video compression \cite{BMVC2016_9,scheerlinck2018continuous,brandli2014real},
event-based feature tracking \cite{kueng2016low,Alzugaray2018asy}. Special pixels are shown in Figure \ref{fig:special_pixel}.

\begin{itemize}
	\item
	\emph{Hot Pixels} are pixels that have lower threshold values than the assumed value. This means that both positive and negative events are generated more often than expected.
	\item 
	\emph{Cold Pixels} are pixels that have higher threshold values than the assumed value. This means that both positive and negative events are generated less often than expected.
	\item 
	\emph{Warm Pixels} are pixels that have negative bias values. This means positive events are triggered more often, and negative events are triggered less often than expected.
	\item    
	\emph{Cool Pixels} are pixels that have positive bias values. This means negative events are triggered more often, and positive events are triggered less often than expected. 
\end{itemize}

\section{Proposed Methods}
\subsection{Offline Calibration: Hybrid Case (Event + Intensity)}

Because each event could cause reset bias 
in the reference log intensity level \cite{yang2015dynamic},  the direct integration with constant contrast threshold 
as in \eqref{1.2} cannot perfectly represent the actual intensity change. 
For DAVIS event cameras, which also produce intensity images as well as event data, we introduce a calibration method that uses both data sources.

Since the biased event-based sensor DVS does not affect the APS circuit which generates intensity images, the corresponding intensity images provide reliable and independent intensity references \cite{brandli2014real}.
Therefore for any pixel $\vb*{p}$, given two log intensity images $L^{\vb*{p}}(T_k)$ and $L^{\vb*{p}}(T_{k+d})$ recorded at times $T_k$ and $T_{k+d}$ respectively, the following approximation holds
\begin{align} \label{1.7}
\sum_{t^{\vb*{p}}_i \in (T_k, T_{k+d}]} \begin{bmatrix} \sigma^{\vb*{p}}_i & |\sigma_i^{\vb*{p}}| \end{bmatrix} \begin{bmatrix} c^{\vb*{p}}(t_i^{\vb*{p}}) \\ b^{\vb*{p}}(t_i^{\vb*{p}}) \end{bmatrix} = \Delta^{k+d}_{k}L^{\vb*{p}},
\end{align}
where $\Delta^{k+d}_{k}L^{\vb*{p}} = L^{\vb*{p}}(T_k) - L^{\vb*{p}}(T_{k+d})$.

Since this approximation holds for any choice of $k$ and $d$, we may stack these equations into the following overdetermined linear system.
For any pixel coordinates $\vb*{p}$, and any choice of $k$, $d$ and $n$,

\begin{equation} \label{simple1.8}
A x = z,
\end{equation}

\begin{equation} \label{1.8}
\begin{split}
\begin{bmatrix} \sum_{t^{\vb*{p}}_i \in (T_k, T_{k+d}]} \begin{bmatrix}
\sigma^{\vb*{p}}_i & |\sigma^{\vb*{p}}_i|
\end{bmatrix} \\
\vdots \\
\sum_{t^{\vb*{p}}_i \in (T_{k+(n-1)d}, T_{k+nd}]} \begin{bmatrix}
\sigma^{\vb*{p}}_i & |\sigma^{\vb*{p}}_i|
\end{bmatrix} \\
\end{bmatrix} 
\begin{bmatrix} c^{\vb*{p}} \\ b^{\vb*{p}} \end{bmatrix} \\
=\begin{bmatrix} \Delta^{k+d}_{k}L^{\vb*{p}}\\.\\.\\ \Delta^{k+nd}_{k+(n-1)d} L^{\vb*{p}} \end{bmatrix},
\end{split}
\end{equation}
where $c^{\vb*{p}}$ and $b^{\vb*{p}}$ are assumed to be approximately constant (slowly time-varying) over the time period between $T_k$ and $T_{k+nd}$.

An ordinary least-squares approach provides a solution to the overdetermined system in \eqref{1.8}.
In our offline calibration model \textit{OffEI}, we estimate bias and contrast threshold for each pixel using raw images from the first frame $k$ to the last frame $k+nd$ of a dataset. The number of sampling frames $d$ may be adjusted based on the performance of the direct integration.

\subsection{Online Calibration: Hybrid Case (Event + Intensity)}
The self-timed reset (STR) event encoding mechanism in event cameras generates new events when the amplified change of log intensity is larger than the threshold of a comparator, which could lose events because of event queueing, comparators comparison and other factors \cite{berner2011event,lichtsteiner2008128,yang2015dynamic}.
\cite{brandli2014real} also shows that different scenes, lighting conditions and object moving speeds may lead to different contrast threshold values.
Hence, we introduce an online calibration method to compute dynamic bias and threshold to adjust changing event rate and environment. 

{
	\begin{figure}[!ht]
		\centering    
		\includegraphics[width=8cm]{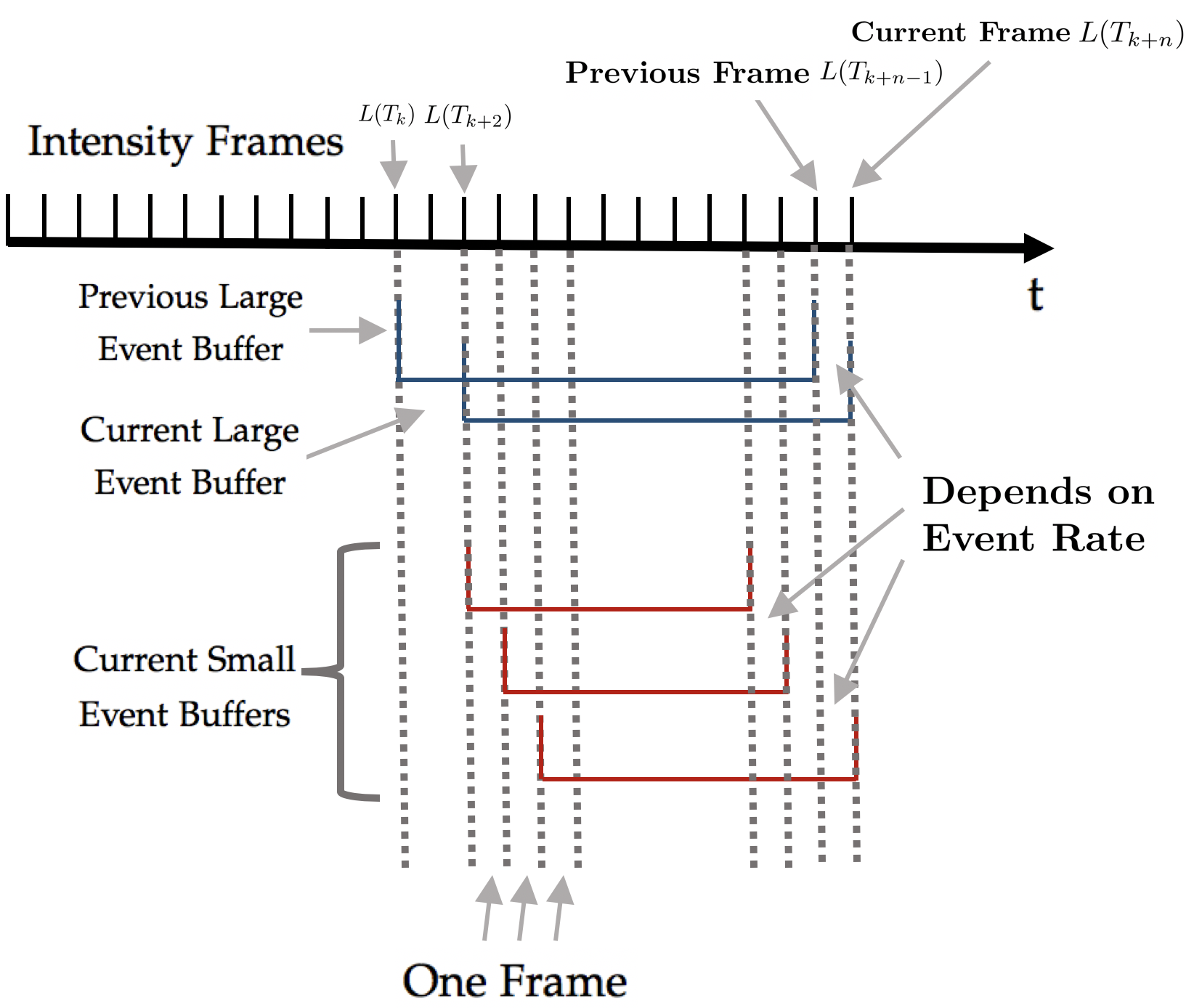}
		\caption{\label{fig:event_buff} Event Buffer Model For Online Calibration Method \textit{OnEI}. Each buffer contains a pre-set number of events and buffer length depends on event rate. The figure shows the current large buffer contains enough events and we start updating $b^{\vb*{p}}$ and $c^{\vb*{p}}$. We divide the large buffer into many small buffers for each least square equation in \eqref{1.8}. Before update, the oldest information between frame $L(T_{k})$ and $L(T_{k+2})$ is removed in the large buffer and the buffer size is decreased by 1 frame. }
	\end{figure}
}

We use a large event buffer and several small event buffers for linear regression in \eqref{1.8}. 
We continuously add new events between $L(T_{k+n-1})$ and $L(T_{k+n})$ 
to the large buffer and increase the buffer length by 1 frame when a new intensity image $L(T_{k+n})$ arrives.
When the large buffer contains enough events (\ie the total number of events in the buffer is larger than a pre-set value), we remove the events that occurred between the oldest image frames $L(T_{k})$ and $L(T_{k+2})$ to decrease the buffer length by 1 frame. Then,
we divide the large buffer into many small buffers. Each small buffer stores a certain amount of events which corresponds to a least square equation. 
We estimate new $b^{\vb*{p}}$ and $c^{\vb*{p}}$ for each pixel using \eqref{1.8}. 
When the large buffer does not contain sufficient events, we will only add new events in the buffer, but not update $b^{\vb*{p}}$ and $c^{\vb*{p}}$ or remove oldest events. 
The time-varying length of two types of event buffer is based on the event rate: shorter buffer for higher event rate and longer buffer for lower event rate. This is illustrated in Figure~\ref{fig:event_buff}.
We assume the bias and contrast threshold are slowly time-varying. Therefore, a low-pass filter is implemented to achieve smoother updates.

\subsection{Offline Calibration: Event-only Case}
We propose a model to estimate contrast threshold $c^{\vb*{p}}$ and bias $b^{\vb*{p}}$ for pure event cameras, \eg DVS cameras, which can only produce event streams. 

Without the intensity image as a reference input, we rely on assumptions about the scene and the resulting event stream:
\begin{itemize}
	\item  The gradient of the mean value of the cumulative sum of the event polarity over time is zero. That is, the intensity value at timestamp $T_{k+d}$ is similar to the intensity value at timestamp $T_k$. 
	\item 
	The variance of the cumulative sum of the event polarity from its mean value is approximately equivalent across all pixels. That is, the scene is roughly equally textured, such that the excitation of each pixel is similar. 
\end{itemize}
These assumptions apply for long sequences of data, while observing an equally textured scene. 

\textbf{Bias $b^{\vb*{p}}$ Calibration:} Compute $b^{\vb*{p}}$ to remove the non-zero gradient, such that, for any pixel coordinates $\vb*{p}$,
\begin{align} \label{1.5}
\mathbb{E}\begin{bmatrix} \sum_{t^{\vb*{p}}_i \in (T_k, T_{k+d}]} \left( c^{\vb*{p}} \sigma^{\vb*{p}}_i + b^{\vb*{p}} \right)\end{bmatrix} = 0.
\end{align}

\textbf{Contrast Threshold $c^{\vb*{p}}$ Calibration:}  Compute $c^{\vb*{p}}$ such that, for any two pixel coordinates $\vb*{p}_1, \vb*{p}_2$,
\begin{align} \label{1.6}
&\text{Var} \begin{bmatrix}
\sum_{t^{\vb*{p}_1}_i \in (T_k, T_{k+d}]} \left( c^{\vb*{p}_1} \sigma^{\vb*{p}_1}_i + b^{\vb*{p}_1} \right)
\end{bmatrix} \notag \\
&\hspace{1cm}=
\text{Var} \begin{bmatrix}
\sum_{t^{\vb*{p}_2}_j \in (T_k, T_{k+d}]} \left( c^{\vb*{p}_2} \sigma^{\vb*{p}_2}_j + b^{\vb*{p}_2} \right)
\end{bmatrix}.
\end{align}

The bias and contrast threshold can be computed in three steps. First, the bias for the pixel $\vb*{p}$ can be estimated from the gradient of the linear regression line through the plot of $\sum \sigma_i^{\vb*{p}}$ versus $\sum |\sigma_i^{\vb*{p}}|$.
That is, a linear regression of the cumulative sum of event polarities compared to the number of events at a given pixel coordinates $\vb*{p}$.
Let the linear regression line satisfies the equation $\sum \sigma_i^{\vb*{p}} = m^{\vb*{p}}( \sum |\sigma_i^{\vb*{p}}| ) + d^{\vb*{p}}$. Second, the contrast threshold can be computed by computing the root mean square error $r^{\vb*{p}}$ of $\sum \sigma^{\vb*{p}}_i$ from the linear regressed line. Finally, the bias $b^{\vb*{p}}$ and contrast threshold $c^{\vb*{p}}$ are
\begin{align}
c^{\vb*{p}} &= \frac{\text{median}(r^{\vb*{p}})}{r^{\vb*{p}}}, \\
b^{\vb*{p}} &= -c^{\vb*{p}} m^{\vb*{p}}.
\end{align}

\section{Experiments}

\noindent
\textbf{Dataset} 

\noindent
We evaluated the proposed calibration methods on the publicly available event-camera dataset of 
\texttt{Dynamic Translation} (indoor environment with moving person, increasing speed, low lighting condition), 
\texttt{Office Spiral} (6-DOF, slow motion, low lighting condition) \cite{mueggler2017event}. In addition, we demonstrated the calibration results on our \texttt{Bright Grass} dataset (unstructured outdoor environment, bright lighting condition), which was recorded using another DAVIS240C event camera in a bright natural scene with both fast and slow motion.
The dataset is available online\footnote{\href{https://github.com/ziweiWWANG/Event-Camera-Calibration}{github.com/ziweiWWANG/Event-Camera-Calibration}
}. 

\noindent
\textbf{Implementation Details}

\noindent
For offline model \textit{OffEI}, we set $d=40$ frames in \eqref{1.8} and used the whole sequence of each dataset to estimate $b^{\vb*{p}}$ and $c^{\vb*{p}}$. For online model \textit{OnEI}, 
experimentally, we found out that when the large buffer contains a maximum of 1,700,000 events and each small buffer stores a maximum of 200,000 events for least square estimation in \eqref{1.8}, our online method performs well for different datasets. The calibration algorithms were implemented in MATLAB. 

\begin{table}[!htbp]
	\caption{Experimental Metrics Between Methods} 
	\label{tab:metrics}
	\small 
	\centering 
	\begin{tabular}{llcr} 
		\toprule[\heavyrulewidth]\toprule[\heavyrulewidth]    
		\texttt{Dynamic Translation} & \textbf{RMSE}& \textbf{PSNR} & \textbf{SSIM}  \\
		\midrule        
		DI& \quad 34.01  & \quad 35.13  & 0.35  \\
		OffE & \quad 26.11 & \quad 39.65 &  0.47  \\
		\cite{brandli2014real} & \quad 34.09 & \quad 35.11 &  0.43 \\
		\textbf{OffEI} & \quad \textbf{22.99} & \quad   \textbf{42.02} &  \textbf{0.55} \\
		\textbf{OnEI} & \quad \textbf{26.23} & \quad  \textbf{39.68} &  \textbf{0.54} \\        
		\midrule
		\texttt{Office Spiral} & \textbf{RMSE}& \textbf{PSNR} & \textbf{SSIM}  \\
		\midrule
		DI & \quad  41.45 & \quad 31.66 & 0.58  \\
		OffE & \quad 35.46 & \quad 34.32  &  0.71  \\
		\cite{brandli2014real} & \quad 32.32 & \quad  35.96 &  0.78 \\
		\textbf{OffEI} & \quad \textbf{27.40} & \quad \textbf{38.77} &  \textbf{0.85}   \\
		\textbf{OnEI} & \quad \textbf{29.79} & \quad  \textbf{37.36} &  \textbf{0.82} \\
		\midrule
		\texttt{Bright Grass} & \textbf{RMSE}& \textbf{PSNR} & \textbf{SSIM}  \\
		\midrule
		DI & \quad  81.16 & \quad  20.02 & 0.36 \\
		OffE & \quad 64.80 & \quad 23.89 & 0.42  \\
		\cite{brandli2014real} & \quad 76.43 & \quad 20.99 &  0.36 \\
		\textbf{OffEI} & \quad \textbf{61.65} & \quad \textbf{24.73} & \textbf{0.50}  \\
		\textbf{OnEI}& \quad \textbf{61.78} & \quad  \textbf{24.70} &  \textbf{0.50} \\
		\bottomrule[\heavyrulewidth] 
	\end{tabular}
	\caption{DI is the direct intergration without calibration. \textit{OffE} and \textit{OffEI} are two offline methods we proposed, event-only and event-image calibration. The Root Mean Square Error (RMSE), Peak Signal-to-Noise Ratio (PSNR) and Structural Similarity Index (SSIM) \cite{wang2004image} are shown. The best resluts in each category are marked in bold.}
\end{table}

\subsection{Experimental Results}
To evaluate our proposed calibration algorithms, we directly integrated events between frame $I_{k-n}$ and $I_{k}$ as the estimated intensity image 
$\hat{I}_k = \exp(\log I_{k-n} + \Sigma^{k}_{i = k-n} 
{{c(\vb*{p}_i)} \sigma_i+b(\vb*{p}_i)})$. 
Then, we compared the integration results $\hat{I}_k$ with the referenced intensity image $I_{k}$, which is produced by a DAVIS240C event camera.
Note that in our event-only method \textit{OffE}, intensity images $I_{k-n}$ are not used to estimate bias and contrast threshold.
We also compared the \textit{OnEI} method with the \cite{brandli2014real} calibration method.

\begin{figure}[h!]
	\centering
	\begin{minipage}{0.05\columnwidth}
		\rotatebox[origin=c]{90}{Dynamic Translation}
		\\[9mm]
		\rotatebox[origin=c]{90}{Office Spiral}
		\\[13mm]
		\rotatebox[origin=c]{90}{Bright Grass}
	\end{minipage}
	\begin{minipage}{0.3\columnwidth}
		\includegraphics[width=0.8\columnwidth]{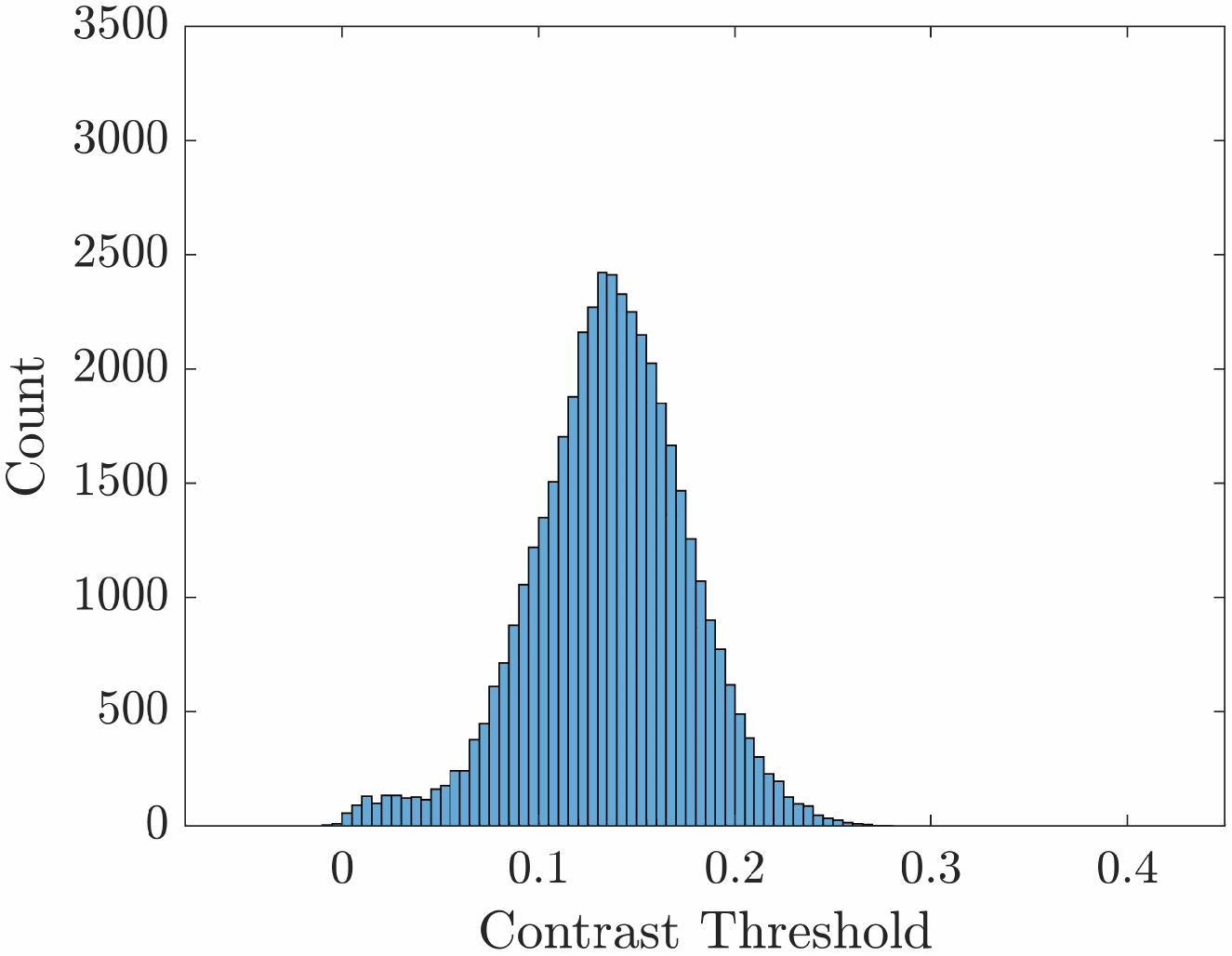}
		\\
		\includegraphics[width=0.8\columnwidth]{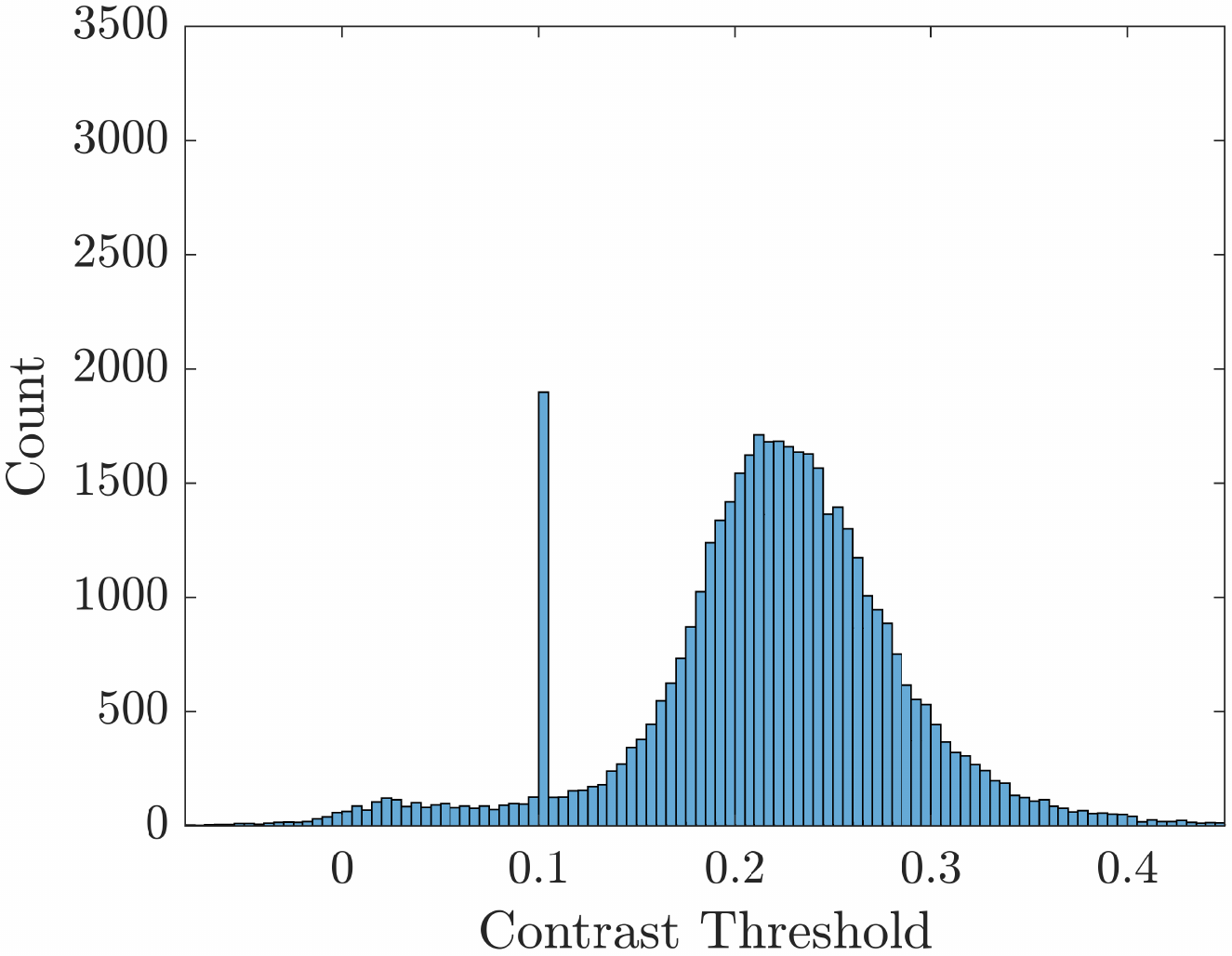}
		\\
		\includegraphics[width=0.8\columnwidth]{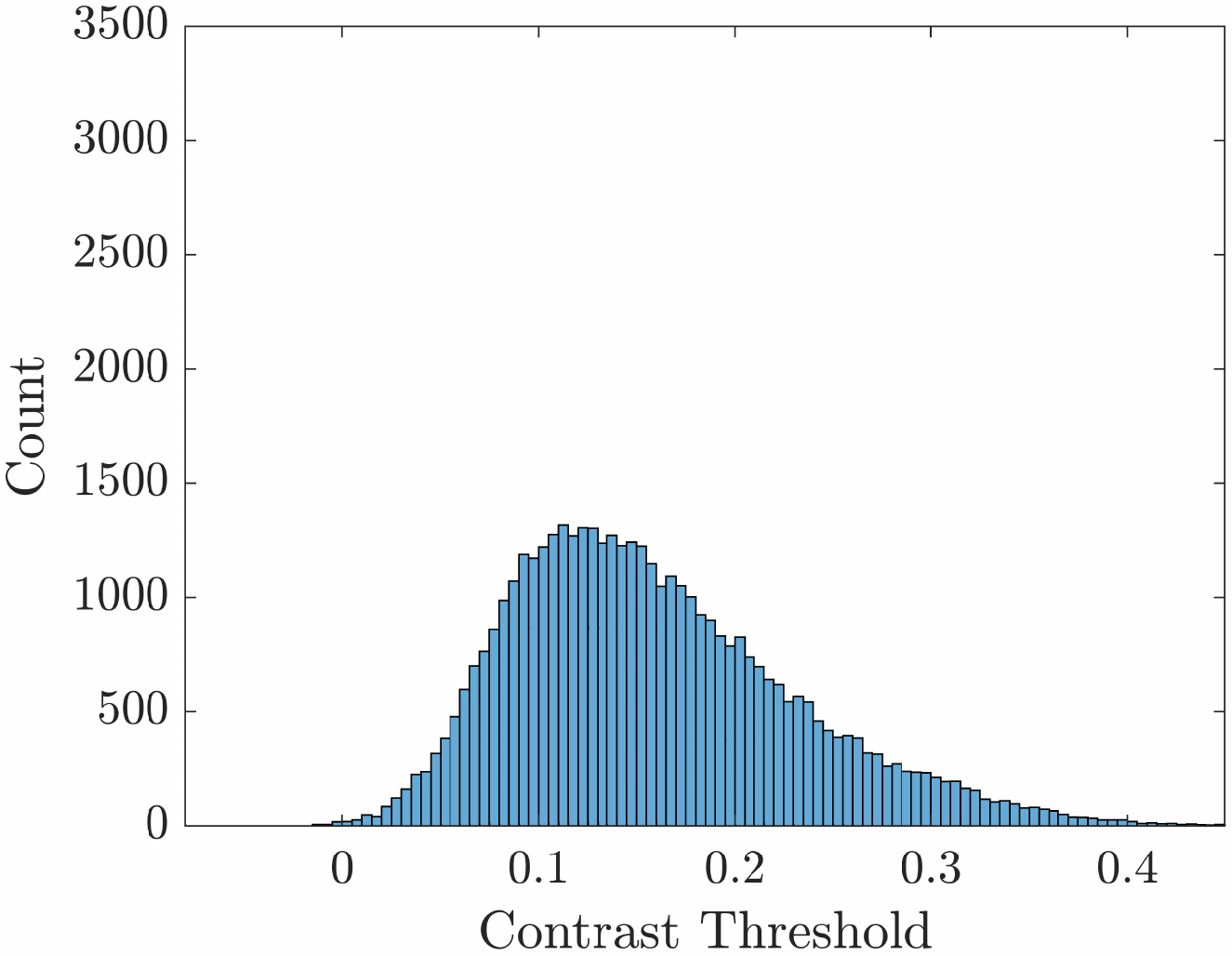}
	\end{minipage}
	\begin{minipage}{0.3\columnwidth}
		\includegraphics[width=0.8\columnwidth]{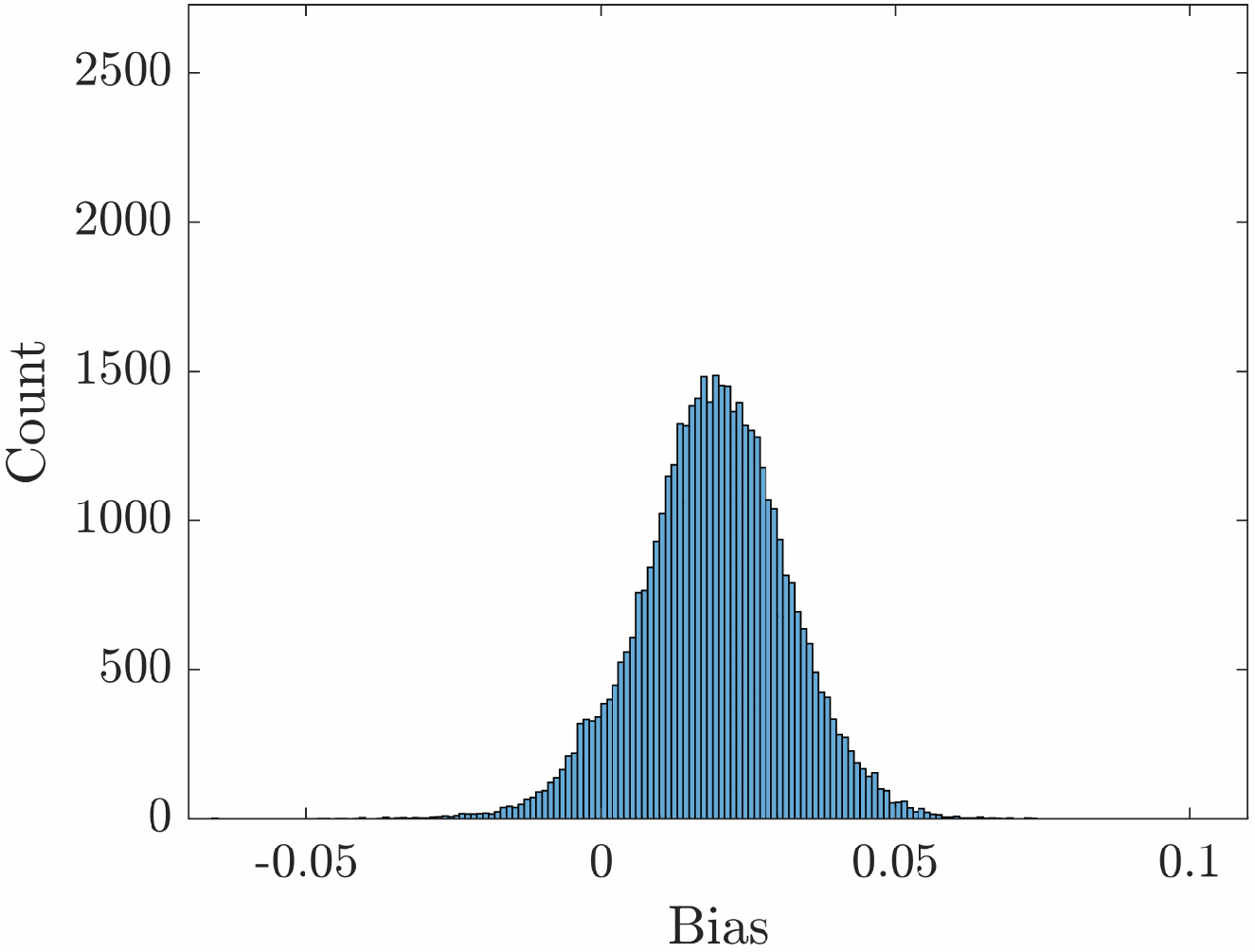}
		\\
		\includegraphics[width=0.8\columnwidth]{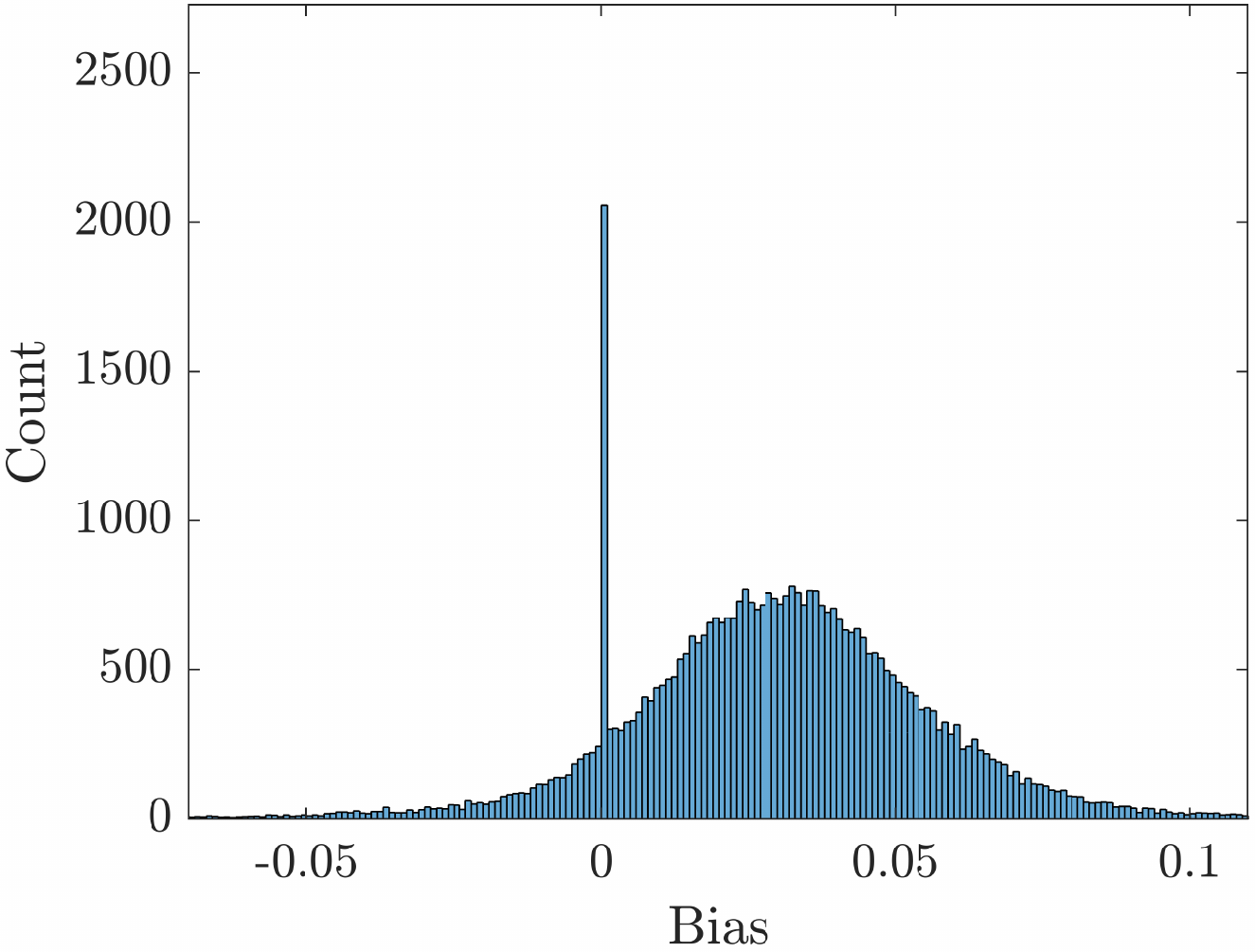}
		\\
		\includegraphics[width=0.8\columnwidth]{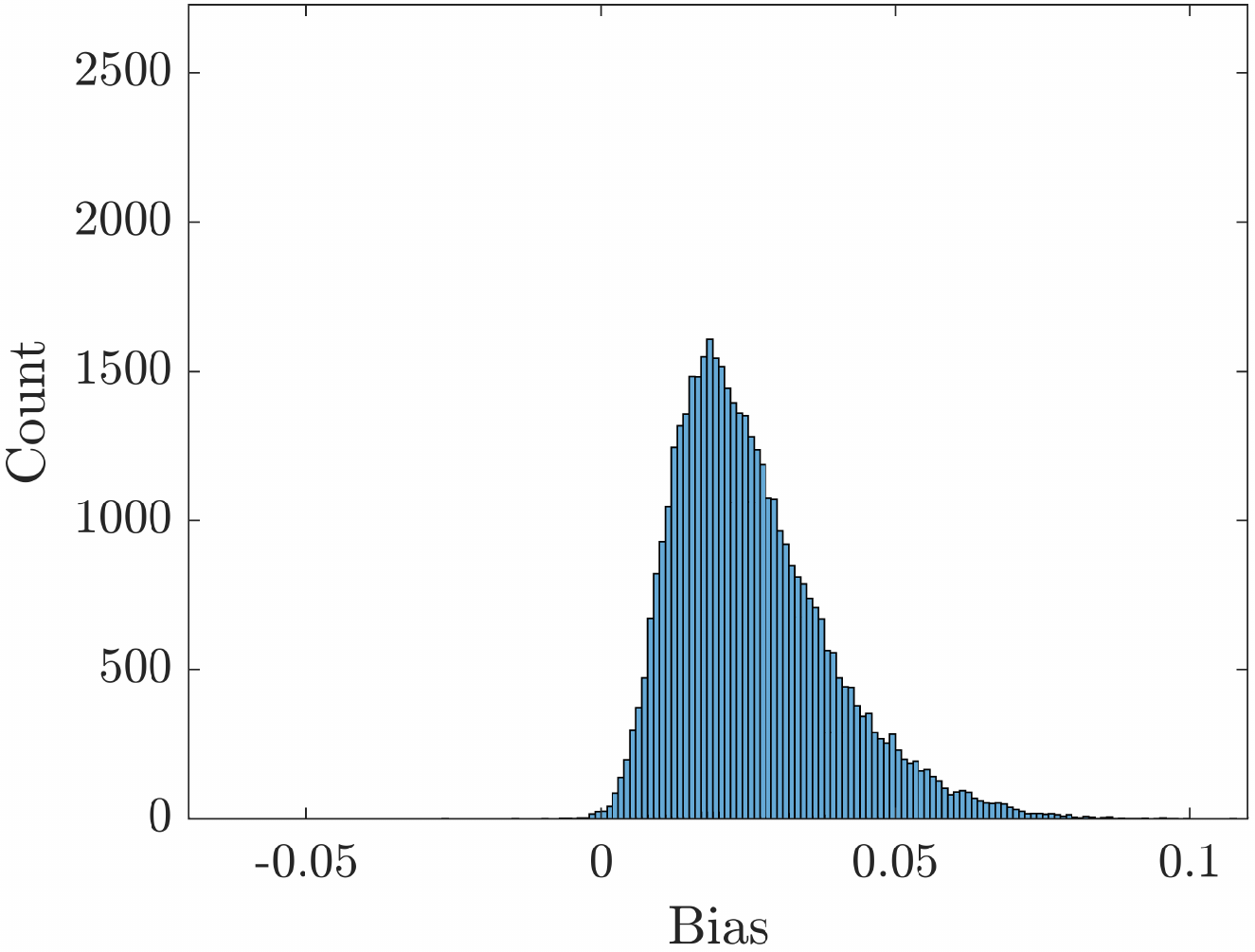}
	\end{minipage}
	\caption{The Histogram of Estimated Bias $b^{\vb*{p}}$ and Contrast Threshold $c^{\vb*{p}}$ Using Different Datasets. First row: \texttt{Dynamic Translation}; second row: \texttt{Office Spiral}; third row:  \texttt{Bright Grass}.}
	\label{fig:hist}
\end{figure}

\begin{figure*}[htbp]
	\centering
	\subfigure[Raw Image]{
		\begin{minipage}[t]{0.3\linewidth}
			\centering
			\includegraphics[width=2in]{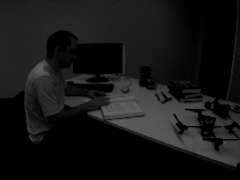}\\
			\vspace{0.01cm}
		\end{minipage}
	}%
	\subfigure[DI]{
		\begin{minipage}[t]{0.3\linewidth}
			\centering
			\includegraphics[width=2in]{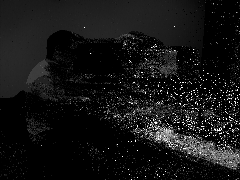}\\
			\vspace{0.01cm}
		\end{minipage}%
	}%
	\subfigure[OffE (ours)]{
		\begin{minipage}[t]{0.3\linewidth}
			\centering
			\includegraphics[width=2in]{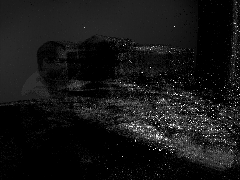}\\
			\vspace{0.01cm}
		\end{minipage}%
	}
	\subfigure[ \protect \cite{brandli2014real}]{
		\begin{minipage}[t]{0.3\linewidth}
			\centering
			\includegraphics[width=2in]{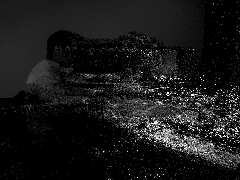}\\
			\vspace{0.01cm}
		\end{minipage}%
	}%
	\subfigure[OnEI (ours)]{
		\begin{minipage}[t]{0.3\linewidth}
			\centering
			\includegraphics[width=2in]{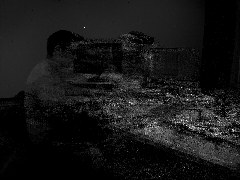}\\
			\vspace{0.01cm}
		\end{minipage}%
	}%
	\subfigure[OffEI (ours)]{
		\begin{minipage}[t]{0.3\linewidth}
			\centering
			\includegraphics[width=2in]{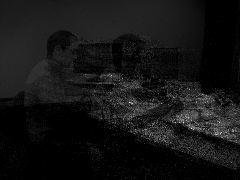}\\
			\vspace{0.01cm}
		\end{minipage}%
	}
	\centering
	\caption{\texttt{Dynamic Translation} Dataset. Raw image shows the frame 366 generated by a standard camera sensor.
		DI is the direct integration without calibration. \textit{OffE} and \textit{OffEI} are two offline methods we proposed, event-only and event-image calibration. We mainly compare our online event-image calibration method OnEI with the \protect \cite{brandli2014real} method. Four calibration methods integrate events between frame 339-366.}
	\vspace{-0.2cm}
	\label{result 2}
\end{figure*}

\begin{figure*}[htbp]
	\centering
	\subfigure[Raw Image]{
		\begin{minipage}[t]{0.3\linewidth}
			\centering
			\includegraphics[width=2in]{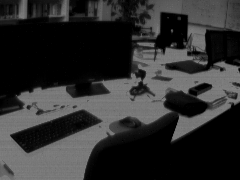}\\
			\vspace{0.01cm}
		\end{minipage}%
	}%
	\subfigure[DI]{
		\begin{minipage}[t]{0.3\linewidth}
			\centering
			\includegraphics[width=2in]{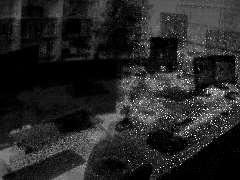}\\
			\vspace{0.01cm}
		\end{minipage}%
	}%
	\subfigure[OffE (ours)]{
		\begin{minipage}[t]{0.3\linewidth}
			\centering
			\includegraphics[width=2in]{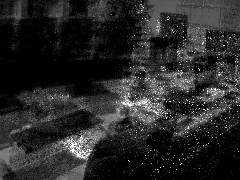}\\
			\vspace{0.01cm}
		\end{minipage}%
	}
	\subfigure[ \protect \cite{brandli2014real}]{
		\begin{minipage}[t]{0.3\linewidth}
			\centering
			\includegraphics[width=2in]{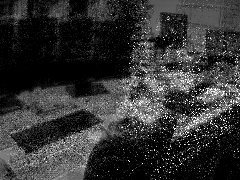}\\
			\vspace{0.01cm}
		\end{minipage}%
	}%
	\subfigure[OnEI (ours)]{
		\begin{minipage}[t]{0.3\linewidth}
			\centering
			\includegraphics[width=2in]{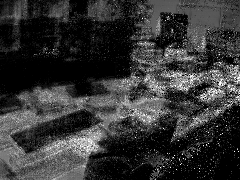}\\
			\vspace{0.01cm}
		\end{minipage}%
	}%
	\subfigure[OffEI (ours)]{
		\begin{minipage}[t]{0.3\linewidth}
			\centering
			\includegraphics[width=2in]{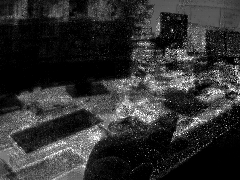}\\
			\vspace{0.01cm}
		\end{minipage}%
	}
	\centering
	\caption{\texttt{Office Spiral} Dataset. Raw image shows the frame 101 generated by a standard camera sensor. DI is the direct intergration without calibration. \textit{OffE} and \textit{OffEI} are two offline methods we proposed, event-only and event-image calibration. We mainly compare our online event-image calibration method \textit{OnEI} with the \protect \cite{brandli2014real} calibration method. Four calibration methods integrate events between frame 83-101.}
	\vspace{-0.2cm}
	\label{result 1}
\end{figure*}

\begin{figure*}[htbp]
	\centering
	\subfigure[Raw Image]{
		\begin{minipage}[t]{0.3\linewidth}
			\centering
			\includegraphics[width=2in]{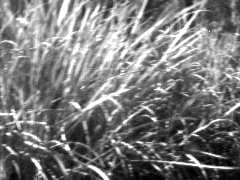}\\
			\vspace{0.01cm}
		\end{minipage}%
	}%
	\subfigure[DI]{
		\begin{minipage}[t]{0.3\linewidth}
			\centering
			\includegraphics[width=2in]{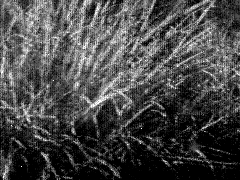}\\
			\vspace{0.01cm}
		\end{minipage}%
	}%
	\subfigure[OffE (ours)]{
		\begin{minipage}[t]{0.3\linewidth}
			\centering
			\includegraphics[width=2in]{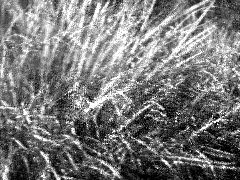}\\
			\vspace{0.01cm}
		\end{minipage}%
	}
	\subfigure[ \protect \cite{brandli2014real}]{
		\begin{minipage}[t]{0.3\linewidth}
			\centering
			\includegraphics[width=2in]{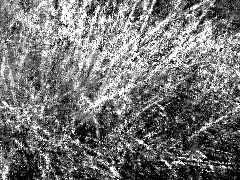}\\
			\vspace{0.01cm}
		\end{minipage}%
	}%
	\subfigure[OnEI (ours)]{
		\begin{minipage}[t]{0.3\linewidth}
			\centering
			\includegraphics[width=2in]{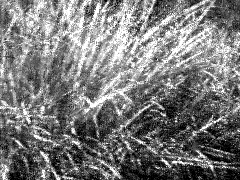}\\
			\vspace{0.01cm}
		\end{minipage}%
	}%
	\subfigure[OffEI (ours)]{
		\begin{minipage}[t]{0.3\linewidth}
			\centering
			\includegraphics[width=2in]{offei_grass.png}\\
			\vspace{0.01cm}
		\end{minipage}%
	}
	\centering
	\caption{\texttt{Bright Grass} Dataset. Raw image shows the frame 225 generated by a standard camera sensor. DI is the direct integration without calibration. \textit{OffE} and \textit{OffEI} are two offline methods we proposed, event-only and event-image calibration. We mainly compare our online event-image calibration method \textit{OnEI} with the \protect \cite{brandli2014real} calibration method. Four calibration methods integrate events between frame 217-225.}
	\vspace{-0.2cm}
	\label{result 3}
\end{figure*}

Exemplary integration results of different methods on three datasets are shown in Figure~\ref{result 2}--\ref{result 3}. The three datasets are recorded under different lighting conditions and motions. 
Because most of the pixels are cool pixel ($b^{\vb*{p}}$ larger than 0), as shown in Figure \ref{fig:hist}, after direct integration without calibration, the Direct Integrated (DI) images will become darker. The hot pixels (white spots) are also obvious in DI images.
Comparing with direct integration using a constant of 0.1 as the contrast threshold, the effect of special pixels is significantly reduced after calibration. For example, in Figure \ref{result 1}, the ``shadow'' left by the previous location of the keyboard in DI image is mostly removed after calibration.
Our intensity image-based method \textit{OffEI} and \textit{OnEI} provide the best calibration results among the four methods presented. For two online methods with intensity images as reference, 
our online method \textit{OnEI} highly reduces the effect of special pixels and noisy background pixels compared to the method in \cite{brandli2014real}, and has higher similarity with the reference intensity image. 
A quantitative comparison of four calibration methods is further discussed in Section \ref{metrics}, where we demonstrate the superiority of our calibration methods.

For our offline method \textit{OffEI}, we also showed the estimated histogram of bias and contrast threshold in Figure \ref{fig:hist}. 
Three different datasets have similar $b^{\vb*{p}}$ and $c^{\vb*{p}}$ distributions. 
The values of bias $b^{\vb*{p}}$ are relatively small, and for most pixels, they are larger than 0, which are consistent with the effect of positive leakage voltage in amplifier \cite{yang2015dynamic}. 
The estimated contrast threshold values $c^{\vb*{p}}$ are between 0 to 0.4. For pixels that do not have any triggered event and where the intensity changes between intensity frames are close to zero during the whole dataset, matrix $A$ is not full rank and most of the values in vector $z$ are close to zero in \eqref{simple1.8}. In addition, the value of $b^{\vb*{p}}$ and $c^{\vb*{p}}$ will not affect the result of the direct integration in these pixels, we directly assign $b^{\vb*{p}}$ to zero and $c^{\vb*{p}}$ to 0.1. For example, some pixels on the monitor screen in \texttt{Office Spiral} remain at zero intensity value during the whole dataset and generate few events (Figure \ref{result 1}). This causes the presence of the two peaks at $c^{\vb*{p}}=0.1$ and $b^{\vb*{p}}=0$ in the second row of Figure \ref{fig:hist}. We also found that the contrast threshold values are not always close to the commonly used value of 0.1 in Figure \ref{fig:hist}. This is likely due to the different camera sensitivity settings used in the three datasets. In a darker environment, a lower sensitivity is usually chosen to keep the events generated at a desired rate.

\subsection{Experimental Metrics} \label{metrics}

With the corrected parameters $b^{\vb*{p}}$ and $c^{\vb*{p}}$, the direct integrated images should be similar to the corresponding intensity images, even though the intensity images generated by DAVIS cameras could be blurry in fast-motion. The Root Mean Square Error (RMSE), Peak Signal-to-Noise Ratio (PSNR) and Structural Similarity Index (SSIM) \cite{wang2004image} are implemented to evaluate the calibration performance. We compare the referenced intensity images from DAVIS camera with the calibrated direct integrated images. Lower RMSE, higher PSNR and SSIM represent higher similarity to the reference image and a better calibration result. 

The average metrics over three different datasets are shown in Table \ref{tab:metrics}. Each output images is direct integrated result for around 500,000 new events from the previous output image. Note that the calibration method in \cite{brandli2014real} and \textit{OnEI} require time to converge, so the evaluation does not include the output images in the first 5 seconds. 
The results are consistent with Figure~\ref{result 2}--\ref{result 3}. Our online method \textit{OnEI} has better calibration performance than the \cite{brandli2014real} method. 
In three datasets, the offline model \textit{OffEI} is slightly better than the online model \textit{OnEI}, which is likely due to the bias and contrast threshold being approximately constant for the datasets we used. 
Another possible reason is that the scenes in the three datasets are relatively unchanged (people and desk, office, grass). Therefore, if the environment and motion speed are known prior to an experiment, it is possible to pre-calibrate the event camera using our offline method. 
Without intensity images, the performance of our event-only method \textit{OffE} is worse than our image-based methods \textit{OnEI} and \textit{OffEI} (except RMSE in \texttt{Dynamic Translation}), which is expected. However,
it is still better than the \cite{brandli2014real} method in \texttt{Dynamic Translation} and \texttt{Bright Grass}.

\section{Conclusion}

To the best of our knowledge, an event generation model supported by transistor junction leakage bias and contrast threshold variance has not been previously proposed.
We fully take advantage of the low-frequency intensity images provided by DAVIS event cameras, providing both online and offline calibration methods.
We also present an offline regression method to calibrate pure event cameras.
We demonstrate that our calibration methods greatly eliminate the effects of special pixels and better reflect the actual intensity changes, which will improve the performance of many applications that use event cameras.

\clearpage

{\small
	\bibliographystyle{ieee_fullname}
	\bibliography{template_arxiv}
}

\end{document}